\documentclass{article}

\usepackage[preprint]{neurips_2021}

\usepackage[utf8]{inputenc} 
\usepackage[T1]{fontenc}    
\usepackage{url}            
\usepackage{booktabs}       
\usepackage{amsfonts}       
\usepackage{nicefrac}       
\usepackage{microtype}      
\usepackage{graphicx}
\usepackage{amsmath}
\usepackage{amsthm}
\usepackage{amssymb}
\usepackage{natbib}
\usepackage{wrapfig}
\usepackage{booktabs}
\usepackage{float}
\usepackage[dvipsnames,table]{xcolor}
\setcitestyle{numbers}

\usepackage[colorlinks = true,
            linkcolor = blue,
            urlcolor  = magenta,
            citecolor = blue,
            anchorcolor = blue,
            backref=page]{hyperref}

\definecolor{lightgray}{rgb}{0.83, 0.83, 0.83}

\newcommand{\gdumbC}{GDumb~\cite{prabhu2020greedy} }

\title{Smaller Is Better: An Analysis of Instance Quantity/Quality Trade-off in Rehearsal-based Continual Learning}

\author{%
  Francesco Pelosin \\
  Dipartimento di Scienze Ambientali, Informatica e Statistica \\
  Ca' Foscari University \\
  Venice, Italy \\
  \texttt{francesco.pelosin@unive.it} \\
  \And
  Andrea Torsello \\
  Dipartimento di Scienze Ambientali, Informatica e Statistica \\
  Ca' Foscari University\\
  Venice, Italy \\
  \texttt{andrea.torsello@unive.it} \\
}

\begin{document}

\maketitle

\begin{abstract}
The design of machines and algorithms capable of learning in a dynamically changing environment has become an increasingly topical problem with the increase of the size and heterogeneity of data available to learning systems.    
As a consequence, the key issue of Continual Learning has become that of addressing the stability-plasticity dilemma of connectionist systems, as they need to adapt their model without forgetting previously acquired knowledge. Within this context, rehearsal-based methods \emph{i.e.}, solutions in where the learner exploits memory to revisit past data, has proven to be very effective, leading to performance at the state-of-the-art.
In our study, we propose an analysis of the memory quantity/quality trade-off adopting various data reduction approaches to increase the number of instances storable in memory. In particular, we investigate complex instance compression techniques such as deep encoders, but also trivial approaches such as image resizing and linear dimensionality reduction.
Our findings suggest that the optimal trade-off is severely skewed toward instance quantity, where rehearsal approaches with several heavily compressed instances easily outperform state-of-the-art approaches with the same amount of memory at their disposal. Further, in high memory configurations, deep approaches extracting spatial structure combined with extreme resizing (of the order of $8\times8$ images) yield the best results, while in memory-constrained configurations where deep approaches cannot be used due to their memory requirement in training, Extreme Learning Machines (ELM) offer a clear advantage.
\end{abstract}

\section{Introduction}
\label{sec:intro}
Continual Learning (CL) is increasingly at the center of attention of the research community due to its promise of adapting to the  dynamically changing environment resulting from the huge increase in size and heterogeneity of data available to learning systems. It has found applications in several domains. Its prime application, and still most active field, is computer vision, and in particular object detection~\cite{gidaris2018dynamic, thrun1996learning,  Parisi2019ContinualLL_review0}; however it has since found applications in several other domains such as segmentation~\cite{cermelli2020modeling, michieli2019incremental, DBLP:journals/corr/abs-2012-03362}, where each segmented class has to be learned in an incremental fashion, as well as in other fields, among which we mention Reinforcement Learning (RL)~\cite{rl0, rl1, rl2} and Natural Language Processing (NLP)~\cite{nlp0, nlp1, nlp2, nlp3}.

Ideally, the behaviour of CL systems should resemble human intelligence in its ability to incrementally learn in a dynamical environment~\cite{Hadsell2020EmbracingCC_review1}, with minimal waste of resources, spatial or computational. The main problem encountered by these systems resides in the famous stability-plasticity dilemma of neuroscience, resulting in the so called {\em catastrophic forgetting}~\cite{MCCLOSKEY1989109}, a phenomenon where new information dislodges or corrupts previously learned knowledge, resulting in the deterioration of the ability to solve previously learned tasks.

Solutions to this problem typically incur in a increase in resource requirements~\cite{lomonaco_cvpr_2020} both for CL's very nature (the more tasks arrive the more data the agent need to process), and for the nature of the systems that try to solve it, both in the increased complexity of the typically  deep learning  models, and in the time and space requirements of continuously learning multiple models. This problem become particularly evident in rehearsal-based methods.

\begin{figure}[t]
    \centering
    \includegraphics[width=1\textwidth]{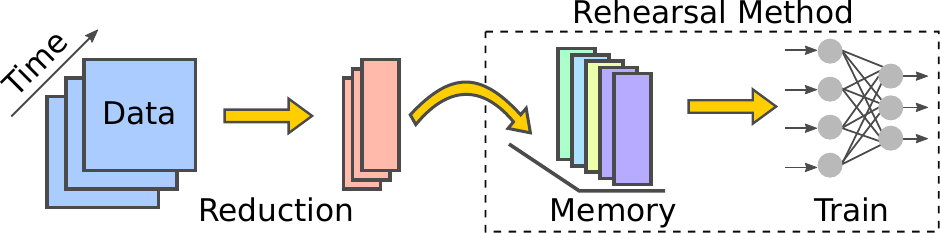}
    \caption{Our work analyzes the optimal instance quantity/quality trade-off in memory buffers of rehearsal-based Continual Learning systems. We carry out our analysis by applying several dimensionality reduction schemes to increase the quantity of storable data.}
    \label{fig:pipeline}
\end{figure}

Rehearsal-based methods, {\em i.e.}, approaches that leverage a memory buffer to cope with catastrophic forgetting, are emerging as the most effective methodology to tackle CL. Their performance, backed by extensive empirical evidence~\cite{lomonaco_cvpr_2020}, finds also a theoretical justification in Knoblauch and co-workers' finding that optimally solving CL would require perfect memory of the past~\cite{pmlr-v119-knoblauch20a}. In fact, if we were able to completely re-train a new system with all previous data every time a new task arrives, Continual Learning  would not appear to be any different from any other learning problem. However, this approach is both spatially and computationally infeasible for most real-world problems and we can argue it is precisely these memory and computational limitations that characterize CL and distinguish it from other learning problems.

Our investigation aims to analyze the trade-offs on limited-memory CL systems. In particular, we focus on the quantity/quality trade-off for memory instances. We do so through the analysis of several dimensionality-reduction schemes applied to data instances that allows us to increase the number of instances storable in our fixed-capacity memory.
In particular we adopted deep learning encoders such as a variation of ResNet18~\cite{resnet18} and Variational Autoencoders (VAE)~\cite{DBLP:journals/corr/KingmaW13}, the simple yet surprisingly effective extreme resizing of image data, and, lastly, we explored Random Projections for dimensionality reduction. The latter scheme turns out to be very effective in low memory scenarios also reducing the model's parameter complexity. Indeed, we will show that a variation of Extreme Learning Machines (ELM) offers a simple yet effective solution for resources-constrained CL systems.

Our analysis will focus on computer vision tasks and use \gdumbC as a rehearsal-baseline. GDumb is a model that has been proposed to question the community's progress in CL thanks to the fact that in lieu of its outstanding simplicity, it was still able to provide state-of-the-art performances. Further, its simplicity also results in high versatility, as it proposes a general CL formulation comprising all task formulations in the literature. GDumb is fully rehearsal-based, and it is composed by a greedy sampler and a dumb learner, that is, the system does not introduce any particular strategy in the selection of replay data. Therefore, it represents the ideal candidate method to carry out our analysis. 

The experimental findings highlighted in our paper are multiple: first, we show that when the memory buffer is fixed and extreme values of resizing of instance data is applied, we can easily push the state-of-the-art of CL rehearsal systems by a minimum of $+6\%$ to a maximum of $+67\%$ in terms of final accuracy. This is a surprising result suggests that the optimal trade-off between data quantity and quality is severely skewed toward the former and that in general the informational content required to correctly classify images in standard datasets is relatively low. Then, we analyze the consumption of resources of rehearsal CL systems as we saturate the rehearsal buffer, and show that ELM offer a clear solution on CL systems constrained by very low resources environments. 

Code and experiments are open and accessible at \href{https://git.io/JG1Yo}{https://git.io/JG1Yo}

\section{Related Work}


Following some recent surveys~\cite{Parisi2019ContinualLL_review0, Hadsell2020EmbracingCC_review1, Mundt2020AWV_review2}, we divide CL approaches into three main categories: regularization-based approaches, data rehearsal-based approaches and architectural-based approaches. Although a few novel theoretical frameworks based on meta-learning have been introduced recently~\cite{Hadsell2020EmbracingCC_review1}, the majority still fall within these categories (or in a mixture of them).

\paragraph{Regularization-based:}
Regularization-based approaches address catastrophic forgetting by controlling each parameter's importance through the subsequent tasks through the addition of a finely-tuned regularizing loss criterion. 
Elastic Weight Consolidation (EWC)~\cite{kirkpatrick2017overcoming} was the first well established approach of this class. It uses Fisher information to estimate each parameter's importance while discouraging the update for parameters with greatest task speciﬁcity. Learn without Forgetting (LwF)~\cite{DBLP:conf/eccv/LiH16} exploits the concept of ``knowledge distillation'' to preserve and regularize the output for old tasks. More recently,  Learning without Memorizing (LwM)~\cite{dhar2019learning} adds in the loss an information preserving penalty exploiting attention maps, Continual Bayesian Neural Networks (UCB)~\cite{ebrahimi2020} adapts the learning rate according to the uncertainty defined in the probability distribution of the weights in the network, while Pomponi \emph{et al.}~\cite{pomponi2020} propose a regularization of network's latent embeddings.

\paragraph{Rehearsal-based:}
 Rehearsal-based solutions allocate a memory buffer of a predefined size and devise some smart schemes to store previously used data to be replayed in the future, {\em i.e.}, to be added to future training samples. One of the first methodologies developed is Experience Replay (ER)~\cite{DBLP:conf/nips/RolnickASLW19}, which stores a small subset of previous samples and uses them to augment the incoming task-data.
Aljundi \emph{et al.}~\cite{aljundi2019online} propose an evolution of ER which takes  in consideration Maximal Interfered Retrieval (ER-MIR). Their proposal lies between rehearsal and regularization methods, its strategy is to retrieve the samples that are most interfered, i.e. whose prediction will be most negatively impacted by the foreseen parameters update. Among other mixed approaches we have Rebuffi \emph{et al.}~\cite{rebuffi2017icarl} proposed a method which simultaneously learns strong classifiers and data representation (iCaRL). Gradient Episodic Memory (GEM)~\cite{gem} and its improved version Averaged-GEM (AGEM)~\cite{agem} exploits the memory buffer to constrain the parameter updates and stores the previous samples as trained points in the parameter space, while Gradient based Sample Selection (GSS) ~\cite{aljundi2019online} diversifies/prioritizes the gradient of the examples stored in the replay memory. Finally, a recent method proposed by Shim \emph{et al.}~\cite{shim2020online} scores memory data samples according to their ability to preserve latent decision boundaries (ASER). 

\paragraph{Architectural-based:}
Architectural methods alter their parameter space for each task. The most influential architectural-based approach is arguably Progressive Networks (PN)~\cite{rusu2016}, where a dedicated network is instantiated for each task while Continual Learning with Adaptive Weights (CLAW)~\cite{adel2020} grows a network that adaptively identifies which parts to share between tasks in a data-driven approach. Note that, in general, the approaches that use incremental modules suffer the lack of task labels at test time, since there is no easy way to decide which module to adopt.

\section{Methodology}

Before introducing the dimensionality reduction approaches adopted in our quantity/quality analysis we have to introduce the CL scenario considered and its task composition. Unfortunately the  community has not yet converged to a unique standard way to define a CL setting~\cite{van2019three}. Here we adopt GDumb's formulation which is the most general one and specifically resembles Lomonaco and Maltoni's formulation~\cite{lomonaco2017core50}. In particular, we focus on the new class (NC)-type scenario~\cite{lomonaco2017core50} where each task $T_i$ introduces data instances of $C$ new, previously unseen, classes. We also consider the more realistic class incremental scenario, that is, we are not allowed to know task labels at test time. 

As incremental approach we use the recently proposed GDumb, which is composed of a simple learner and a greedy balancer. That is, given a fixed amount of memory $\mathcal{M}$, each instance of task data is randomly sampled in order to balance class instances in the memory, so that, at the end of the $T_i$ task experience, the memory contains an equal number of instances of all previously encountered classes i.e. each class has $\left \lfloor \frac{\mathcal{M}}{C * i} \right \rfloor$ instances in memory.  

Besides providing state of the art performances, GDumb has been proposed as standard baseline to question our progresses in continual learning research, since after experiencing a task, the simple learner (such as a ResNet18~\cite{resnet18} or a MLP) is trained \emph{only} with memory data, making GDumb a fully rehearsal based approach with random filtering of incoming data, and thus the ideal candidate to carry our study. In the following paragraphs, we briefly describe all the strategies adopted for dimensionality reduction.  

\begin{figure}[t]
    \centering
    \includegraphics[width=1\textwidth]{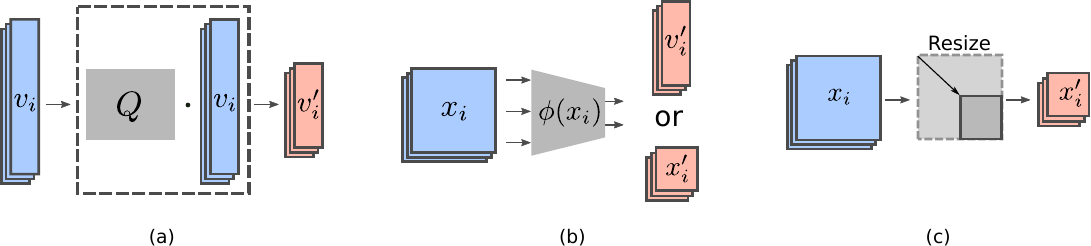}
    \caption{Depiction of the three main dimensionality reduction techniques analyzed. In  (a) random projection (RP) each image is vectorized ($v_i$) and then orthogonally-projected through a random matrix $Q$ into the compressed representation $v^\prime_i$. In (b)  encoding strategy, the encoder $\phi$ outputs a latent vector  $v_i^\prime$ (such as in VAEs) or a noise-free / shrinked image $x_i^\prime$ (as in CutR). In (c) we adopt a simple image resizing strategy through standard biliniear interpolation.}
    \label{fig:techniques}
\end{figure}
\subsection{Random Projections (RP)}
Extreme Learning Machines (ELM)~\cite{huang_universal_2006} are a set of algorithms that exploit random projections as dimensionality reduction technique to preserve computational and spatial resources while learning. ELM  have been introduced in 2006 and recently have found application in neuroscience~\cite{elm_neuro0, elm_neuro1, elm_neuro2} and in other problems such as in molecular biology~\cite{elm_chem0}. The idea can be roughly described as a composition of two modules where the first one performs a random projection of the data, while the second one is a learning model. The appealing property of RP lies in the Johnson-Lindenstrauss lemma~\cite{Johnson1984ExtensionsOL} which states that given a set of points in a high dimensional plane, there is a linear map to a subspace that roughly preserves the distances between data points by some approximation factor.



The Johnson-Lindenstrauss lemma guarantees that we can obtain a low-distortion to the dimensionality reduction by multiplying each instance vector by a semi-orthogonal random matrix $Q^{m \times n}$ in the $(m,n)$ Stiefel manifold. More formally, let $x_i$ be an image of the current task of width, height and number of channels $w$, $h$, and $c$ respectively, then the size of $x_i$ is $n = h w c$.  We can consider its vectorization as $v_{i} \in \mathbb{R}^{n}$ and its compressed representation 

\begin{equation}
v_{i}^{\prime} = Q v_{i}  \hspace{15pt} \text{s.t.} \hspace{15pt} Q^{T} Q = I_{m}
\end{equation}
with $v_{i}^{\prime} \in \mathbb{R}^{m}$.

The usage of ELM unsuspectedly unlocks two main advantages: First it allows us to exploit the dimensionality reduction by \emph{increasing the number data instances} storable in the memory buffer. Secondly and, more importantly, allows us to \emph{use models with significantly fewer parameters}. On the other hand, the approach loses coordinate contiguity and, with that, shift co-variance, rendering convolutional approaches inapplicable.

After the random projection data instances will be forwarded to the greedy sampler of GDumb to fill the memory $\mathcal{M}$. Then we perform a rehearsal train with any MLP-like architecture, resulting in an order-of-magnitude reduction in the amount the amount of parameters needed to process visual data allowing the usage of CL rehearsal based solutions in very low resource scenarios.


\subsection{Deep Encoders} 
Deep encoders are neural models $\phi$ that take as input an image $x_{i}$ and, depending from the structure of such model, can output either a latent vectorial representation $v_{i}^{\prime}$, or a squared feature map which we consider as a noise-free shrinked image $x_i^\prime$. Figure \ref{fig:techniques} (b) reports visually the two possible encoding scenarios. In this work, we adopt a Variational AutoEncoder (VAE)~\cite{DBLP:journals/corr/KingmaW13} for the first case and a pretrained ResNet18~\cite{resnet18} cut up to a predefined block (CutR) as a prototype for the second.

\paragraph{VAE} Variational Autoencoders~\cite{DBLP:journals/corr/KingmaW13} have been introduced as an efficient approximation of the posterior for arbitrary probabilistic models. A VAE is essentially an autoencoder that is trained with a reconstruction error between the input and decoded data, with a surplus loss that constitutes a variational objective term attempting to impose a normal latent space distribution. The variational loss is typically computed through a Kullback-Leibler divergence between the latent space distribution and the standard Gaussian, the total loss can be summarized as follows:
\begin{equation}
    \mathcal{L}=\mathcal{L}_{r}(x_{i},\hat{x_{i}}) +\mathcal{L}_{KL}(q(z_{i}|x_{i}), p(z_{i}))
\end{equation}

given an input data image $x_{i}$, the conditional distribution $q(z_{i}|x_{i})$ of the encoder, the standard Gaussian distribution $p(z_{i})$, and the reconstructed data $\hat{x_{i}}$. We use the encoding part of a VAE pretrained on a predefined dataset by feeding each incoming image and retrieving the vectorial output representation $v_i^\prime$, then the greedy sampler filter it for the memory $\mathcal{M}$. 

\paragraph{CutR} As our second encoding approach, we use a pretrained ResNet18~\cite{resnet18} cut up to a predefined block.
ResNets models are convolutional neural networks (CNNs) introducing skip connections between convolutional blocks to alleviate the so called vanishing gradient~\cite{hochreiter1998vanishing} problem afflicting deep architectures. 
The idea behind it, is to use the cut ResNet18 as a \emph{filtering module} that outputs a smaller feature map, giving us $x_{i}^{\prime}$. In fact, we cut the network towards later blocks, since, neurons in the last layers, encode more structured semantics with respect to the early ones~\cite{olah2017feature}. Therefore we are able to extract semantic knowledge from unseen images leveraging transfer learning~\cite{tan2018survey}, that is, we exploit the ability of a model to generalize over unseed data. We refer to this method with the name CutR(esnet18). We use CutR instance encoding by feeding each image belonging to the current task and retrieving the shrinked output $x_i^\prime$ which is then forwarded to the greedy sampler module of GDumb to fill the memory $\mathcal{M}$.

In our analysis, we adopted the less resource-hungry VAE scheme  for datasets where shift co-variance is not as important, such as the MNIST, in which the digits are centered in the image and thus most approaches at the state-of-the-art use a MLP as classifier. In all other instances we used the CutR scheme.

\subsection{Resizing}
We used also the simplest instance reduction approach one can think of {\em i.e.},  resizing  the images to very low resolution through standard bilinear interpolation. The resized images are then fed to the sampler of GDumb to balance the classes in $\mathcal{M}$ and all  training and prediction is performed on the  lowered resolution images.

Independently of the approach adopted, all data instances are reduced before storing them in  memory, then we use GDumb's greedy sampler to select and balance class instances, and finally , we use a suitable learner of to fit memory data and assess the performance. 
In general, following GDumb, we adopt ResNet18 for large-scale image classification tasks for all approaches that maintain shift co-variance, reverting to a simple MLP for approaches without shift co-variance like RP.


\section{Experiments}
\label{sec:experiments}
We performed our analysis on the following  standard  benchmarks:
\begin{itemize}
\item MNIST~\cite{mnist}: the dataset is composed by 70000 $28\times28$ grayscale images of handwritten digits divided into 60000 training  and 10000 test images belonging to 10 different classes.
\item CIFAR10~\cite{cifar}: consists of 60000 RGB images of objects and animals. The size of each image is $32\times32$ divided in $10$ classes, with 6000 images per class. The dataset is split into 50000 training images and 10000 test images. 
\item CIFAR100~\cite{cifar}: is composed by 60000, $32\times32$ RGB images subdivided in 100 classess with 600 images each. The dataset is split into 60000 training images and 10000 test images. 
\item ImageNet100~\cite{imagenet}: the dataset is composed of $64\times64$ RGB images divided in 100 classes; it is composed of  60000 images split into 50000 training and 10000 test. 
\item Core50~\cite{lomonaco2017core50}: the dataset is composed of $128\times128$ RGB images of domestic objects divided in 50 classes. The set consists of 164866 images split into  115366  training and 49500 test. 
\end{itemize}

Following~\cite{prabhu2020greedy}, we use final accuracy as the evaluation metric throughout the paper.
The metric is computed \textit{at the end of all tasks} against a test set of never seen before images  composed of an equal number of instances per  class. This allows us to directly compare against the largest number of competitors in the literature.


All the experiments has been conducted with an Intel i7-4790K CPU with 32GB RAM and a 4GB GeForce GTX 980 machine running \texttt{PyTorch 1.8.1+cu102}.

\begin{figure}[t]
    \centering
    \includegraphics[width=1\textwidth]{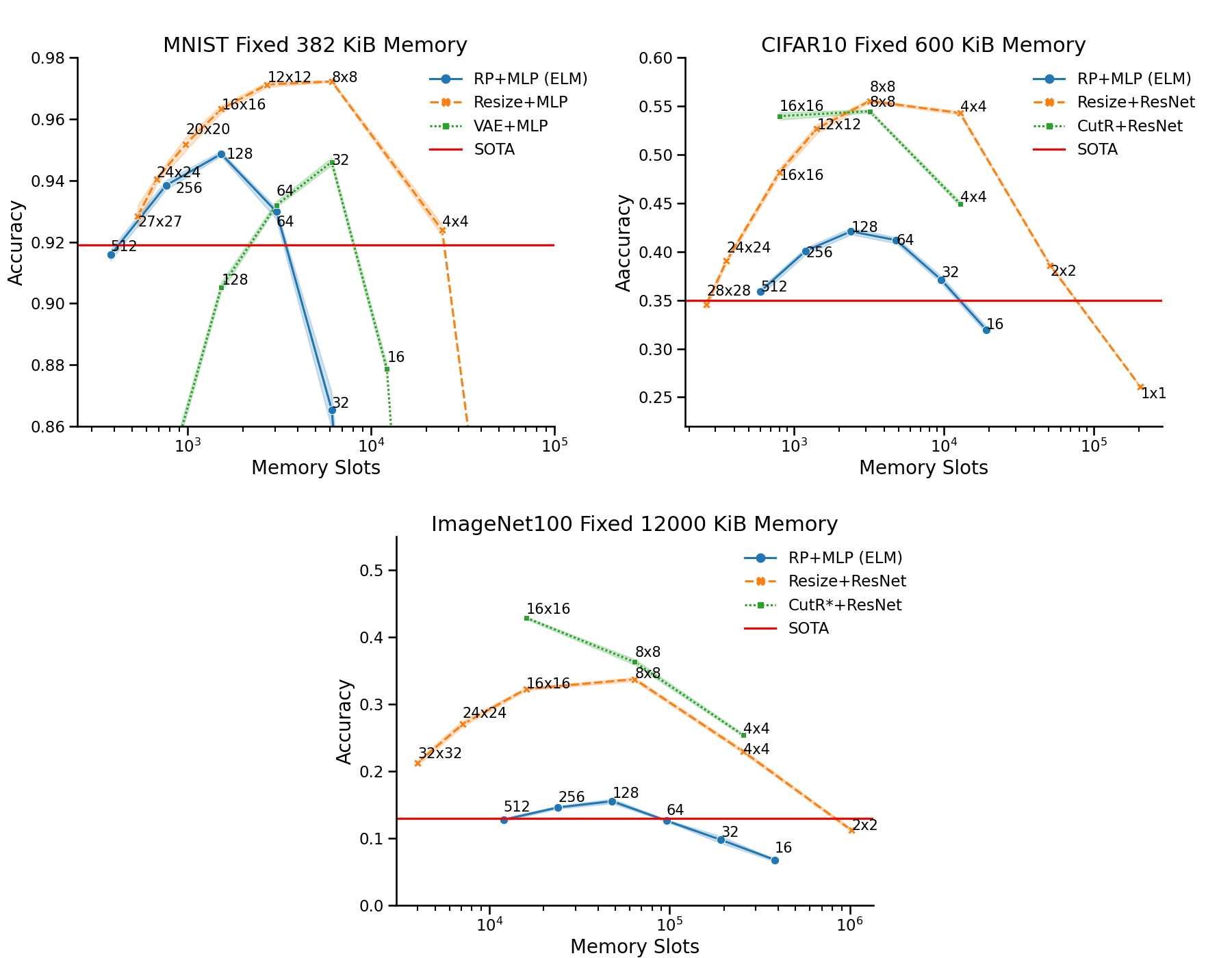}
    \caption{At top-left the accuracy analysis of the MNIST dataset. In top-right we have the analysis of CIFAR10 and at bottom we have ImageNet100. The state-of-the-art (SOTA) method is plain GDumb with an MLP as incremental learner in the MNIST experiment and Resnet18 in the others. The number of instances in memory (i.e. the $x$ axis) is in $log$ scale. We report the results of (5 runs).}
    \label{fig:optimal}
\end{figure}

\subsection{Parameter Sensitivity}

In the first experiment, we compared different dimensionality reduction strategies as we altered the parameters. The analysis was conducted on three different datasets: MNIST, CIFAR10 and ImageNet100.  In this evaluation we fixed the amount of memory buffer used for GDumb during rehearsal training, and we measured the final accuracy as the parameters varied for each dimensionality reduction method. In particular we subdivided both MNIST and CIFAR10 datasets into 5 tasks of 2 classes each,  with 600 KiB dedicated memory buffer, while ImageNet100 was divided into 10 tasks of 10 classes each, with 12 MiB memory buffer.  
 
Figure~\ref{fig:optimal} plots the performance of the various  schemes as we reduce the dimensionality of the instances and and thus increase their number in the allocated memory.  
The orange line represents the performance of the resize scheme. For the MNIST dataset, we considered eight different target sizes\footnote{throughout the paper we omit to write the channel component for brevity} $x_{i}^{\prime} \in \{27\times27, 24\times24, 20\times20, 16\times16, 12\times12, 8\times8, 4\times4, 2\times2, 1\times1\}$. We performed the same resizing for CIFAR10 data. We did not report CIFAR100 analysis since the data format is the same as CIFAR10 and the result would be analogous. For ImageNet100, we resized each instance to $x_{i}^{\prime} \in \{32\times32, 24\times24, 16\times16, 6\times6, 4\times4, 2\times2\}$. 


The green line of Figure~\ref{fig:optimal} represents the  deep encoders. In particular, for MNIST  we used a VAE \cite{DBLP:journals/corr/KingmaW13} pretrained on KMNIST \cite{clanuwat2018deep} and analyzed the performances of GDumb with compressed instances as we altered the size of the latent embedding vector to  $v_{i}^{\prime} \in \{128, 64, 32, 16\}$. On the other hand, for the CIFAR10  and ImageNet100 dataset we considered different parameters for CutR. In particular, we cut the ResNet18 up to the sixth layer to get a $4\times4$ output, to the fifth to have a $8\times8$ encoding, and lastly up to the third block to get a $16\times16$ feature map. 

\begin{wraptable}{l}{7cm}
\scriptsize
\centering

\begin{tabular}{@{}cccc@{}}
\toprule
\multicolumn{4}{c}{\textbf{CIFAR10}} \\
\textit{Method} & \textit{Acc@600KiB} & \textit{Acc@1.5MiB} & \textit{Acc@3MiB} \\
\midrule
EWC \cite{kirkpatrick2017overcoming} & 17.9 $\pm$ 0.3 & 17.9 $\pm$ 0.3 & 17.9 $\pm$ 0.3 \\
GEM \cite{gem} & 16.8 $\pm$ 1.1 & 17.1 $\pm$ 1.0 & 17.5 $\pm$ 1.6 \\
AGEM \cite{agem}                       & 22.7 $\pm$ 1.8 & 22.7 $\pm$ 1.9 & 22.6 $\pm$ 0.7 \\ 
iCARL \cite{rebuffi2017icarl}          & 28.6 $\pm$ 1.2 & 33.7 $\pm$ 1.6 & 32.4 $\pm$ 2.1 \\
ER \cite{DBLP:conf/nips/RolnickASLW19} & 27.5 $\pm$ 1.2 & 33.1 $\pm$ 1.7 & 41.3 $\pm$ 1.9 \\
ER-MIR \cite{aljundi2019online}        & 29.8 $\pm$ 1.1 & 40.0 $\pm$ 1.1 & 47.6 $\pm$ 1.1 \\
ER5    \cite{aljundi2019online}        & -              &  -             & 42.4 $\pm$ 1.1 \\
ER-MIR5 \cite{aljundi2019online}       & -              &  -             & 49.3 $\pm$ 0.1 \\
GSS \cite{gss}                         & 26.9 $\pm$ 1.2 & 30.7 $\pm$ 1.2 & 40.1 $\pm$ 1.4 \\
ASER \cite{shim2020online}             & 27.8 $\pm$ 1.0 & 36.2 $\pm$ 1.1 & 43.1 $\pm$ 1.2 \\
ASER$_\mu$ \cite{shim2020online}       & 26.4 $\pm$ 1.5 & 36.3 $\pm$ 1.2 & 43.5 $\pm$ 1.4 \\
\rowcolor{red!15}
GDumb \cite{prabhu2020greedy}          & 35.0 $\pm$ 0.6 & 45.8 $\pm$ 0.9 & 61.3 $\pm$ 1.7 \\
\midrule
\rowcolor{orange!15}
Resize ($8\times8$) & \textbf{55.5 $\pm$ 0.2} & \textbf{64.5 $\pm$ 0.2} & \textbf{73.1 $\pm$ 0.2} \\
\rowcolor{blue!15}
ELM ($128$)    & \textbf{43.0 $\pm$ 0.3} & \textbf{47.1 $\pm$ 0.2} & 50.0 $\pm$ 0.2          \\
\rowcolor{green!15}
CutR ($8\times8$)   & \textbf{54.4 $\pm$ 0.2} & \textbf{60.9 $\pm$ 0.2} & \textbf{71.6 $\pm$ 0.6} \\ 
\bottomrule
\end{tabular}

\caption{CIFAR10 final accuracy (5 runs) analysis as we vary the memory for all schemes considered.}
\label{tbl:cifar10}
\vspace{-1em}
\end{wraptable}

The CutR Resnet18 has been pretrained on the complete ImageNet, thus the results in the ImageNet100 benchmark can be biased. We denote these biased results with CutR\textbf{*}.


\begin{table}[t]
\scriptsize%
\begin{minipage}{.6\textwidth}

  \centering

\begin{tabular}{@{}ccccc@{}}
\toprule
\multicolumn{1}{c|}{} & \multicolumn{2}{c|}{\textbf{ImageNet100}} & \multicolumn{2}{c|}{\textbf{CIFAR100}} \\
\textit{Method} & \textit{Acc@12MiB} & \textit{Acc@24MiB} & \textit{Acc@3MiB} & \textit{Acc@6MiB} \\
\midrule
AGEM \cite{agem} & 7.0 $\pm$ 0.4 & 7.1 $\pm$ 0.5 & 9.05 $\pm$ 0.4 & 9.3 $\pm$ 0.4 \\
ER \cite{DBLP:conf/nips/RolnickASLW19} & 8.7 $\pm$ 0.4 & 11.8 $\pm$ 0.9 & 11.02 $\pm$ 0.4 & 14.6 $\pm$ 0.4 \\
EWC \cite{kirkpatrick2017overcoming} & 3.2 $\pm$ 0.3 & 3.1 $\pm$ 0.3 & 4.8 $\pm$ 0.2 & 4.8 $\pm$ 0.2 \\
GSS  \cite{gss} & 7.5 $\pm$ 0.5 & 10.7 $\pm$ 0.8 & 9.3 $\pm$ 0.2 & 10.9 $\pm$ 0.3 \\
ER-MIR \cite{aljundi2019online} & 8.1 $\pm$ 0.3 & 11.2 $\pm$ 0.7 & 11.2 $\pm$ 0.3 & 14.1 $\pm$ 0.2 \\
ASER \cite{shim2020online} & 11.7 $\pm$ 0.7 & 14.4 $\pm$ 0.4 & 12.3 $\pm$ 0.4 & 14.7 $\pm$ 0.7 \\
ASER$_{\mu}$ \cite{shim2020online} & 12.2 $\pm$ 0.8 & 14.8 $\pm$ 1.1 & 14.0 $\pm$ 0.4 & 17.2 $\pm$ 0.5 \\
\rowcolor{red!20}
GDumb \cite{prabhu2020greedy} & 13.0 $\pm$ 0.3 & 21.6 $\pm$ 0.3 & 17.1 $\pm$ 0.2 & 25.7 $\pm$ 0.7 \\
\bottomrule
\rowcolor{orange!15}
Resize ($8\times8$) & \textbf{33.6 $\pm$ 0.2} & \textbf{33.6 $\pm$ 0.3} & \textbf{38.5 $\pm$ 0.4} & \textbf{45.1 $\pm$ 0.2} \\
\rowcolor{blue!15}
ELM ($128$) & \textbf{13.3 $\pm$ 0.2} & 15.4 $\pm$ 0.4 & \textbf{22.4 $\pm$ 0.3} & \textbf{25.7 $\pm$ 0.3} \\
\rowcolor{green!15}
CutR ($8\times8$) & \textbf{36.25 $\pm$ 0.4*} & \textbf{36.27 $\pm$ 0.5*} & \textbf{32.6 $\pm$ 0.6} & \textbf{37.1 $\pm$ 0.2} \\
\bottomrule
\end{tabular}

\end{minipage}
\begin{minipage}{.4\textwidth}
  \scriptsize
  \centering
   
\begin{tabular}{@{}cc@{}}
\toprule
\multicolumn{2}{c}{\textbf{MNIST}} \\
\textit{Method} & \textit{Acc@382KiB} \\ \midrule
GEN \cite{DBLP:journals/corr/abs-1810-12488} & 75.5 $\pm$ 1.3 \\
GEN-MIR \cite{aljundi2019online} & 81.6 $\pm$ 0.9 \\
ER \cite{DBLP:conf/nips/RolnickASLW19} & 82.1 $\pm$ 1.5 \\
GEM \cite{gem} & 86.3 $\pm$ 1.4 \\
ER-MIR \cite{aljundi2019online} & 87.6 $\pm$ 0.7 \\
\rowcolor{red!15}
GDumb \cite{prabhu2020greedy} & 91.9 $\pm$ 0.5 \\
\bottomrule
\rowcolor{orange!15}
Resize ($8\times8$) & \textbf{97.2 $\pm$ 0.1} \\
\rowcolor{blue!15}
ELM ($128$) & \textbf{95.0 $\pm$ 0.4} \\
\rowcolor{green!15}
VAE ($32$) & \textbf{94.6 $\pm$ 0.1}
\\ \bottomrule
\end{tabular}

\end{minipage}
\vspace{1em}
\caption{ImageNet100 and MNIST final accuracy (5 runs) analysis as we vary the memory.
\label{tbl:cifarimagenet}}
\vspace{-1em}
\end{table}

Lastly, the blue line of Figure~\ref{fig:optimal} reports the accuracy of  Random Projection followed by an MLP classifier. We recall that this kind of architecture is a variation of an Extreme Learning Machine (ELM), therefore we will refer to it with the term ELM. We analyzed the final accuracy as the size of the random projection changes, in particular the embedding sizes considered are $v_{i}^{\prime} \in \{512, 256, 128, 64, 32, 16\}$ for all the datasets.

For all the experiments in MNIST data, we used a 2-layer MLP with 400 hidden nodes as learning module, while we used a Resnet18~\cite{resnet18} for all the other analysis with exception of ELM scheme that maintains the 2-layer MLP model throughout. We did not perform any hyperparameter tuning on the learning module in accordance with the \gdumbC experimental protocol.   For completeness we report the learning parameters: the system uses an SGD optimizer, a fixed batch size of $16$, learning rates $[0.05,0.0005]$, an SGDR~\cite{DBLP:conf/iclr/LoshchilovH17} schedule with $T_{0}= 1$, $T_{mult}= 2$ and warm start of 1 epoch. Early stopping with patience of 1 cycle of SGDR, along with standard data augmentation is used (normalization of data). GDumb uses cutmix~\cite{DBLP:conf/iccv/YunHCOYC19} with $p=0.5$ and $\alpha=1.0$ for regularization on all datasets except MNIST.

As we can also see from Figure~\ref{fig:optimal} all the strategies considered unlock performances greatly above the state-of-the-art, thus suggesting that the quantity/quality trade-off is severely skewed toward quantity since each dimensionality reduction technique greatly improves the amount of data instances that can be stored in the memory buffer. It is also evident that the simple resizing strategy gives the best performances improving the state-of-the-art by $+6\%$ on MNIST and roughly by $+20\%$ on both CIFAR10 and ImageNet100 datasets.



Moreover, we chose to consider extreme levels of encoding. We did so to find the level of compression that irreversibly corrupts spatial information and thus makes learning impossible. Surprisingly, it turns out that a $2\times2$ resizing still works on CIFAR10 data with perfomances above the state-of-the-art while a $1\times1$ resize is still better than a random classifier whose performance would be $20\%$ of final accuracy. 
This is a strong evidence that the amount of data storable in the memory buffer plays a central role, but also that CIFAR10 dataset constitutes an unrealistic benchmark and should not been considered to assess novel methodologies in the future.

After choosing and fixing the optimal parameters for each compression scheme, we study the performances of the rehearsal system as we alter the quantity of the memory allocated. In Tables~\ref{tbl:cifar10},\ref{tbl:cifarimagenet} we compute the final accuracy for all the datasets previously considered, with the addition of CIFAR100 with an increase of 20\% in performances. The amount of dedicated memory for the rehearsal buffer, has been chosen in order to be consistent with several other methods at the state-of-the-art, allowing us to compare GDumb's performance on optimized memory schemes against other methods. As we can see, all memory optimizations still provide huge advantages as the memory buffer varies, suggesting again, that instance quantity plays a fundamental role in rehearsal systems even with extreme encoding settings.

Finally, one could argue  that the great improvement in performance is symptomatic of the fact that benchmarks are too simplistic and do not represent real world challenges. On the other hand, the deep models used for classification have a large number of degrees of freedom and require a large amount of instances to be properly trained to capture the complexity of the task at hand. Simpler, lower dimensionality instances allow both for more instances and simpler classifiers with fewer parameters without losing  lot of informational content.

\begin{figure}[t]
    \centering
    \includegraphics[width=1\textwidth]{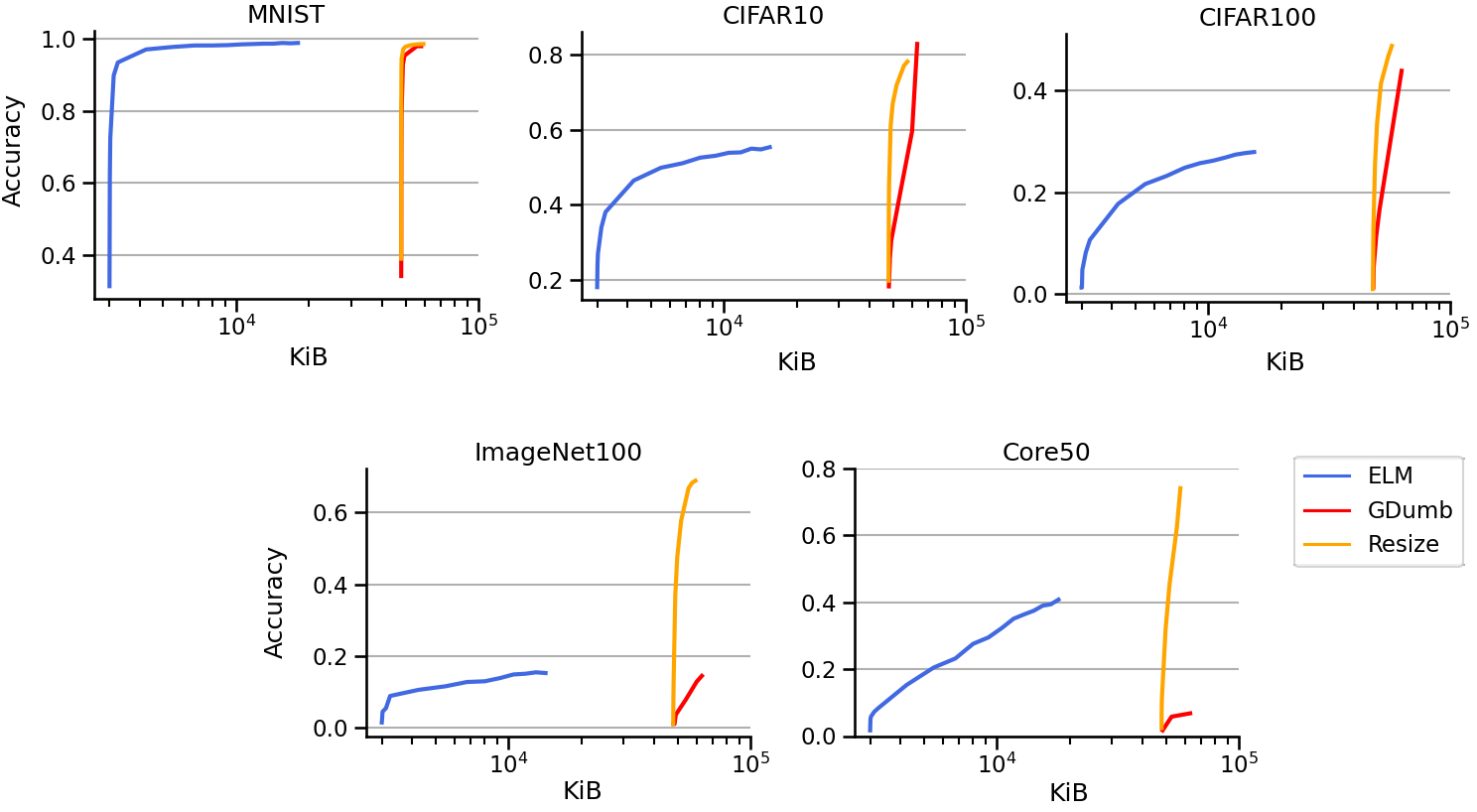} 
    \caption{We show the total amount of KiB used by the whole CL system. We measure the consumption as we saturate the rehearsal memory plus the storage of model parameters. The $x-$axis is in $log$ scale.}
    \label{fig:saturation}
\end{figure}

\subsection{Resource Consumption}

With the second experiment, we wanted to analyze the performance versus the total memory requirement for each approach. Here, we increased the number of instances in the memory buffer and added to the total consumption the working memory used by the classifier to store (and train) the parameters.
We considered three different scenarios: first we used the plain GDumb CL system without dimensionality reduction (representing the state-of-the-art), then we used ELM (with fixed embedding size of ($v_{i}^\prime = 128$), and lastly the resizing scheme (images resized to $x_{i}^\prime = 8\times8$). We selected the best parameters resulting from the previous experiment. 

We then assessed the performance and resource usage using a new dataset, namely the Core50~\cite{lomonaco2017core50}. The reason behind the use of Core50 to validate our findings is twofold: first, we test again whether the quantity of extremely encoded data plays a central role on our rehearsal scheme. Secondly, we measure the performances and the resource usage of a CL system on a more complex set of tasks. We divided the dataset into 10 tasks of 5 classes each.


In Figure~\ref{fig:saturation}, we report the results of this experiment. We can see that extreme levels of resizing still provide optimal results in all the datasets considered. One striking finding is that in Core50 with extreme resizing schemes, even if the size was not optimized for the dataset, the final accuracy is increased by $+67\%$ with respect to the state-of-the-art. Second, we note that ELM constitute a viable solution in low resources scenarios. Indeed, we can surpass the performance of state-of-the-art for low memory scenarios where  even just the classifier used in other approaches could not fit in the allocated memory, much less the rehearsal buffer.
This is clearly observed from the Core50 results. 
We can appreciate that by randomly projecting image data and learning in a low resource scenario provides a boost of $+34\%$ in the final accuracy.

Finally, it is worth noting there is a striking dissonance in the literature of rehersal-based method when the narrative around  buffer-memory sizes revolves around decisions among sizes of the order of 300KiB to 600KiB when then the same systems adopt complex classifiers using several megabytes of memory just for the learned parameters and in the order of gigabytes of working memory for learning. In a real constrained-memory scenario a simpler classifier with more instances offers a clear advantage.

\section{Conclusions}

In this study we analyzed the quantity/quality trade-off in rehersal-based Continual Learning systems adopting 
several dimensionality reduction schemes to increase the number of instances in memory at the cost of possible loss in information. In particular, we used deep encoders, random projections, and a simple resizing scheme. What we found is that even simple, but extremely compressed encodings of instance data provide a notable boost in performance with respect to the state of the art, suggesting that in order to cope with catastrophic forgetting, the optimization of the memory buffer can play a central role. Notably, the performance boost of extreme instance compression suggests that the quality/quantity trade-off is severely biased toward data quantity over data quality.
We suspect that some fault might be in the overly simplistic datasets adopted by the community, but mostly the deep models used for classification are well known to be data-hungry and the instances stored are not sufficent to properly train them, but can suffice for simpler classifiers with fewer parameters working on simplified instances.

Finally, it is worth noting there is a striking dissonance in the literature of rehersal-based method when the narrative around  buffer-memory sizes revolves around decisions among sizes of the order of 300KiB to 600KiB when then the same systems adopt complex classifiers using several megabytes of memory just for the learned parameters and in the order of gigabytes of working memory for learning. In a real constrained-memory scenario a simpler classifier with more instances offers a clear advantage.

Finally, in a real constrained-memory scenario deep convolutional systems using several megabytes of memory for the model parameters and gigabytes of working memory for learning are not a viable solution. On the other hand,  Extreme Learning Machines offer a simple and effective solution for these scenarios.

\bibliographystyle{IEEEtran}
\bibliography{biblio}

\appendix
\section{Other Experiments}

\subsection{Fixed Data Instances}

\begin{figure}[h]
    \centering
    \includegraphics[width=1\textwidth]{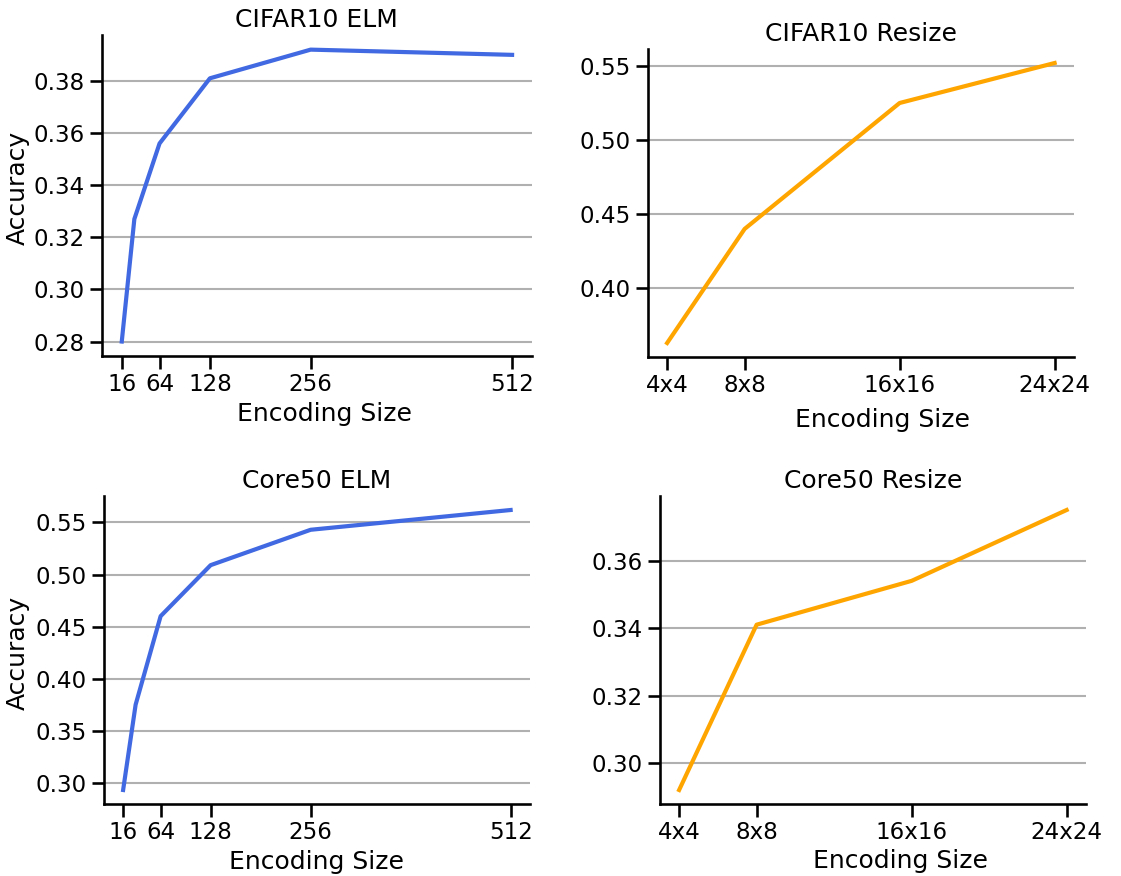}
    \caption{Performance as we vary the parameters for each scheme on CIFAR10 and Core50. In the former benchmark, the memory buffer is of 1000 fixed instances, while in the latter is of 8000.}
    \label{fig:instance}
\end{figure}

With this experiment we aim to better show that instance quantity is preferable over instance quality. We fixed the number of data slots in the memory buffer, and we analyzed the performance as we alter the encoding size. In particular, we tested two datasets namely CIFAR10 and Core50. In CIFAR10 we fixed the buffer to 1000 data slots, while in the latter benchmark we fixed it to be 8000 slots. What we can see from Figure~\ref{fig:instance} is that the improvement of performance is not given by the encoding's smoothing property, and, again, we confirm that rehearsal systems are skewed towards data quantity.

\newpage

\subsection{ELM Width Analysis}
\begin{figure}[h]
    \centering
    \includegraphics[width=0.7\textwidth]{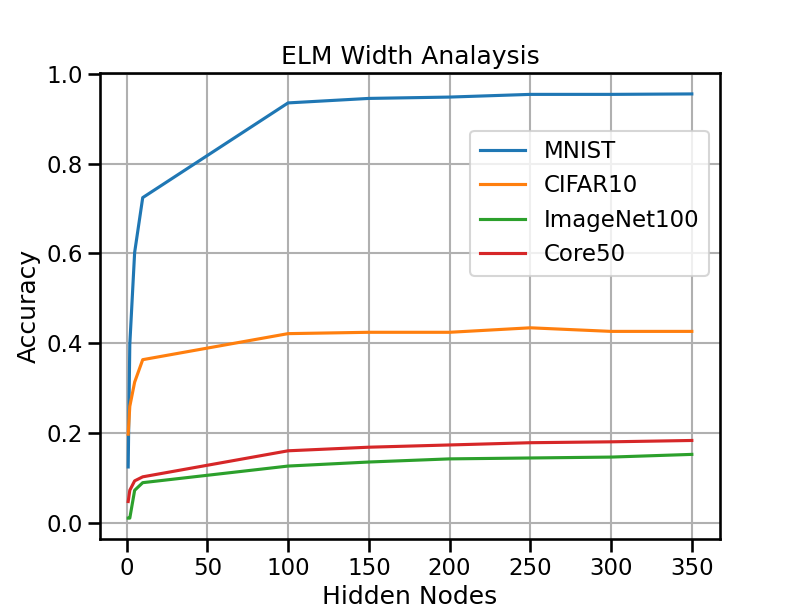}
    \caption{Analysis of final accuracy as we alter the number of hidden units in ELM.}
    \label{fig:width}
\end{figure}

As we specified in the paper, we used a variation of an Extreme Learning Machine. In particular, the architecture is composed by a random projection module and a learning module. The first is implemented through an orthogonal random matrix. While the second is a two layer MLP. Throughout the paper we used 400 hidden units as last layer before the output. We choose to do so to be consistent with GDumb experimental settings. With this experiment we analyze the accuracy metric as we change the number of hidden units. We fixed the encoded size of data to be $v_{i}^{\prime}128$. As memory buffer, we used a different number of data slots for different datasets. That is, for MNIST and CIFAR10 we adopted 2400 slots (600 KiB), in ImageNet100 we used 48000 instances i.e. 12 MiB, while for Core50 we used 8000 slots (2 MiB). In Figure~\ref{fig:width} we can see that 100 hidden units are sufficient to achieve the maximum performance. This, again, shows that more deep classifiers which are common in CL rehearsal literature, might need more data to be trained properly.  

\subsection{Experiments with other Rehearsal Systems}
Throughout our paper, we used GDumb to carry out our analysis. Although we extensively motivated this choice, we also tested two different rehearsal systems. In particular we studied ER~\cite{DBLP:conf/nips/RolnickASLW19} and ER-MIR~\cite{aljundi2019online} performance as we adapt them to work in a low resource scenario. We simply substitute the original learner with our ELM proposal. In Table~\ref{tbl:er} we report the performance of CIFAR10 with 600 KiB buffer memory and $v_{i}^{\prime} = 128$ encoding. As validation metrics we used the final accuracy and the average forgetting~\cite{chaudhry2018riemannian} (lower is better). In order to train the systems, we used the official implementations found at \href{https://github.com/optimass/Maximally_Interfered_Retrieval}{https://github.com/optimass/Maximally\_Interfered\_Retrieval} without any alteration of training hyperparameters. 
As we can see, the results suggest again that ELMs constitute a valid solution for low resource CL systems and that rehearsal solutions are biased toward data quantity over data quality.

\begin{table}[h]
\centering
\begin{tabular}{@{}ccccc@{}}
\toprule
\multicolumn{5}{c}{\textbf{CIFAR10 Fixed Memory 600 KiB}}                    \\
\textit{Method}  & \textit{Accuracy (A)}  & \textit{Forgetting (F)} & \textit{ELM (A)} & \textit{ELM (F)}  \\ \midrule
ER \cite{DBLP:conf/nips/RolnickASLW19}  & 27.5 $\pm$ 1.20  & 48.0 $\pm$ 0.40 & \textbf{42.0 $\pm$ 0.10} &  \textbf{41.2 $\pm$ 0.16} \\
ER-MIR \cite{aljundi2019online} & 29.8 $\pm$ 1.10  & 44.6 $\pm$ 0.48 & \textbf{45.6 $\pm$ 0.10} & \textbf{31.6 $\pm$ 0.01} \\ 
\bottomrule
\end{tabular}

\vspace{10pt}
\caption{Experiments in CIFAR10 with two different rehearsal systems in low resource scenario. }
\label{tbl:er}
\end{table}

\section{Other Specifications}

\subsection{Resource Consumption}

In Table~\ref{tab:summary} we report some summary statistics. In particular, we report GDumb's performance improvements for two encoding schemes i.e. Resize ($8 \times 8$) and ELM ($v_{i}^{\prime}= 128$). We reported only the accuracy according to optimal parameters. We also added the compression factor $\mathcal{C}$, the requirements to store model's parameters $\Theta$ and the memory buffer $\mathcal{M}$. We also report the quantity of GPU memory usage to train GDumb for each encoding scheme. We can see that there is a big gap on the training requirements and memory buffers.  

\vspace{20pt}

\begin{table}[h]
\scriptsize
    \centering

\begin{tabular}{@{}ccccccccc@{}}
\toprule
\multicolumn{1}{l}{}  & \textbf{MNIST} & \textbf{CIFAR10} & \textbf{CIFAR100} & \textbf{ImageNet100} & \textbf{Core50} & \textbf{$\mathcal{C}$} & \textbf{ $\Theta + \mathcal{M}$} & \textbf{GPU} \\ \midrule
\textbf{Resize ($8\times8$)} & (+6\%)         & (+21\%)          & (+20\%)           & (+20\%)              & (+67\%)         & 253:1                & 60 MiB                            & 2.2 GiB           \\
\textbf{ELM ($128$)}    & (+10\%)        & (+10\%)          & (+10\%)           & (+10\%)              & (+10\%)         & 192:1                & 16 MiB                            & 0.72 GiB          \\ \bottomrule
\end{tabular}
    \vspace{10pt}
    \caption{Performance summary and memory consumption}
    \label{tab:summary}
\end{table}



\subsection{Datasets Specification}

For completeness, we report in Table~\ref{tab:memstats} some specifications for the considered datasets. In particular, we provide the task subdivision for each dataset. As we can see MNIST and CIFAR10 have been split in 5 tasks of 2 classes each. This splitting is also known in literature as Split-CIFAR10 and Split-MNIST. For CIFAR10 and ImageNet100 benchmarks we used 10 tasks of 10 classes each, meanwhile for Core50 we shuffled all scenarios and created 10 tasks of 5 classes each. The majority of the works fix the memory slots to define the memory buffer. In our case we used memory requirements expressed in KiB or MiB so that we could alter each slot consumption. We provide a correspondence between memory requirements and  memory slots in the case we consider original image sizes, we do so to ease future comparisons against our work.

\vspace{20pt}

\begin{table}[h]
    \centering
    
\begin{tabular}{@{}ccccc@{}}
\toprule
\multicolumn{5}{c}{\textbf{Experimental Settings}} \\
\textit{Dataset} & \textit{Image size} & \textit{Memory Size} & \textit{Data Slots}  & \textit{Task Composition} \\ \midrule
MNIST & 28x28x1 & 382 KiB & 500 & 5 tasks, 2 classes \\ \midrule
CIFAR10 & 32x32x3 & 600 KiB & 200 & 5 tasks, 2 classes \\
  & & 1.5 MiB & 500 & \\
  & & 3 MiB & 1000 & \\ 
  & & 6 MiB & 2000 & \\
CIFAR100 & - & - & - & 10 tasks, 10 classes \\ \midrule
ImageNet100 & 64x64x3 & 12 MiB & 1000 & 10 tasks, 10 classes\\
  & & 24 MiB & 2000 & \\ \midrule 
Core50 & 128x128x3 & 15 MiB & 312 & 10 tasks, 5 classes \\  
  
  \bottomrule
\end{tabular}

    \vspace{10pt}
    \caption{Dataset and memory statistics, in CIFAR100 row we omit the 2nd, 3rd and 4th columns since are equal to CIFAR10 row.}
    \label{tab:memstats}
\end{table}






\end{document}